\documentclass[runningheads]{llncs}

\usepackage{eccv}

\usepackage{eccvabbrv}

\usepackage{graphicx}
\usepackage{booktabs}
\usepackage{wrapfig}
\usepackage{multirow}
\usepackage{makecell}
\usepackage{amssymb}
\usepackage{pifont}
\usepackage{color} 
\usepackage[dvipsnames]{xcolor}
\usepackage[accsupp]{axessibility}  

\usepackage[pagebackref,breaklinks,colorlinks]{hyperref}

\usepackage{orcidlink}

\begin{document}

\title{Absolute-Unified Multi-Class Anomaly Detection via Class-Agnostic Distribution Alignment} 

\titlerunning{CADA: Absolute-Unified Multi-Class Anomaly Detection}

\author{Jia Guo\inst{1} \and Haonan Han\inst{2}  \and Shuai Lu\inst{2} \and Weihang Zhang\inst{2} \and
Huiqi Li\inst{1}}

\authorrunning{G.~Author et al.}

\institute{School of Information \& Electronics, Beijing Institute of Technology, Beijing, China \and
School of Medical Technology, Beijing Institute of Technology, Beijing, China \\
\email{huiqili@bit.edu.cn}}

\maketitle

\begin{abstract}
Conventional unsupervised anomaly detection (UAD) methods build separate models for each object category. Recent studies have proposed to train a unified model for multiple classes, namely \textit{model-unified} UAD. However, such methods still implement the unified model separately on each class during inference with respective anomaly decision thresholds, which hinders their application when the image categories are entirely unavailable. In this work, we present a simple yet powerful method to address multi-class anomaly detection without any class information, namely \textit{absolute-unified} UAD. We target the crux of prior works in this challenging setting: different objects have mismatched anomaly score distributions. We propose \textbf{C}lass-\textbf{A}gnostic \textbf{D}istribution \textbf{A}lignment (CADA) to align the mismatched score distribution of each implicit class without knowing class information, which enables unified anomaly detection for all classes and samples. The essence of CADA is to predict each class's score distribution of normal samples given any image, normal or anomalous, of this class. As a general component, CADA can activate the potential of nearly all UAD methods under absolute-unified setting. Our approach is extensively evaluated under the proposed setting on two popular UAD benchmark datasets, MVTec~AD and VisA, where we exceed previous state-of-the-art by a large margin.
\end{abstract}

\section{Introduction}
\label{sec:intro}

Anomaly detection aims to detect abnormal patterns from normal images and further localize the anomalous regions. Because of the diversity of potential anomalies and their scarcity, a common practice is to model the accessible training sets containing only normal samples as an unsupervised paradigm, i.e., unsupervised anomaly detection (UAD). UAD has a wide range of applications, e.g., industrial defect detection \cite{bergmann2019mvtec}, video surveillance \cite{tudor2017unmasking}, and medical disease screening \cite{zhao2021anomaly}, addressing the difficulty of collecting and labeling all possible anomalies in these scenarios.

Prior efforts on UAD attempt to learn the distribution of available normal samples. Reconstruction-based methods \cite{shi2021unsupervised,you2022unified,Deng2022,akcay2019ganomaly,zavrtanik2021reconstruction}, such as AutoEncoders \cite{collin2021improved,zavrtanik2021reconstruction} and GANs \cite{akcay2019ganomaly,zhao2021anomaly}, are optimized to reconstruct the images \cite{zavrtanik2021reconstruction, akcay2019ganomaly} or pre-trained features \cite{you2022unified, shi2021unsupervised} of normal images so that anomalies can be detected by reconstruction error. Such methods assume that the networks trained on normal images can reconstruct anomaly-free regions well, but cannot precisely reconstruct anomalous regions \cite{you2022unified,Deng2022}. Distillation-based (or teacher-student) methods \cite{salehi2021multiresolution,cao2022informative,wang2021student} adopt a similar assumption, training a student network from scratch to mimic the anomaly-free features of a pre-trained teacher network. Pseudo-anomaly methods \cite{zavrtanik2021draem,li2021cutpaste} generate pseudo defects or noises on normal images to imitate anomalies, converting UAD to supervised classification \cite{li2021cutpaste} or segmentation tasks \cite{zavrtanik2021draem}. Embedding-statistic methods \cite{defard2021padim, roth2022towards, lee2022cfa} detect anomalies based on the dissimilarity between the features of test images and those of all normal images in the training set (or their reduced selection) extracted from a large-scale pre-trained network.

\begin{figure*}[!t]
\centering
\centerline{\includegraphics[width=\textwidth]{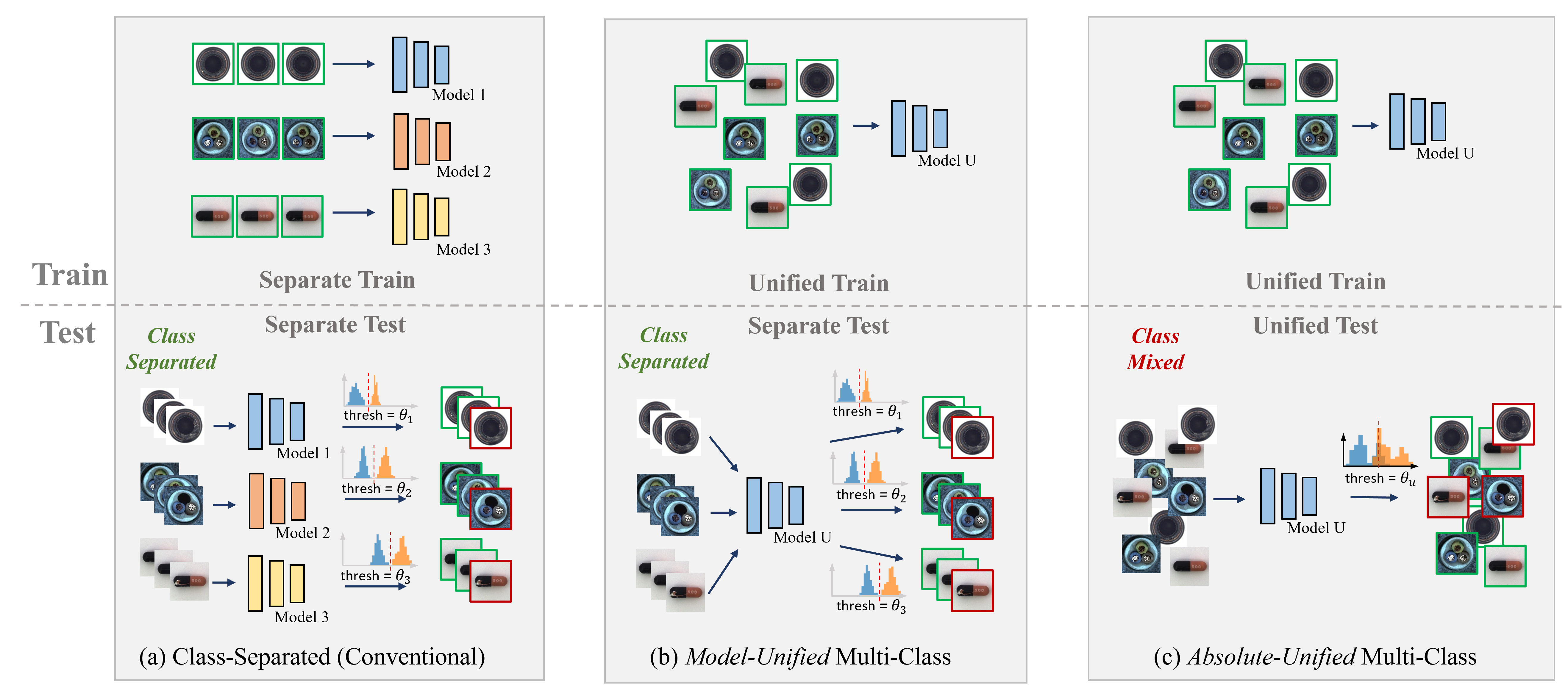}}
\caption{Task settings of (a) class-separated (conventional) UAD, (b) \textit{model-unified} multi-class UAD \cite{you2022unified}, and (c) \textit{absolute-unified} multi-class UAD (proposed).}
\label{fig:1}
\end{figure*}

Most existing UAD methods build a separate model for each object category, as in \cref{fig:1}a. However, this one-class-one-model setting entails substantial storage overhead for saving models \cite{you2022unified}, especially when the application scenario necessitates a large number of object classes. For most UAD methods, A compact boundary of normal patterns is vital to distinguish anomalies. Once the intra-normal patterns become exceedingly complicated due to various classes, the corresponding distribution becomes challenging to measure, consequently harming the detection performance \cite{you2022unified}. Recently, UniAD \cite{you2022unified} and successive studies \cite{guo2023recontrast, yin2023lafite,liu2023mixed} have proposed to train a unified model for multi-class anomaly detection, as illustrated in \cref{fig:1}b. Under this setting, the "identity shortcut" that directly copies the input as the output regardless of normal or anomaly results in the failure of conventional methods \cite{you2022unified}. This phenomenon is caused by the diversity of multi-class normal patterns that drive the network to generalize on unseen patterns. Therefore, strategies such as masked transformer \cite{you2022unified}, contrastive reconstruction \cite{guo2023recontrast}, and diffusion model \cite{yin2023lafite} are proposed to mitigate the identity shortcut.

\begin{figure*}[!t]
\centering
\centerline{\includegraphics[width=\textwidth]{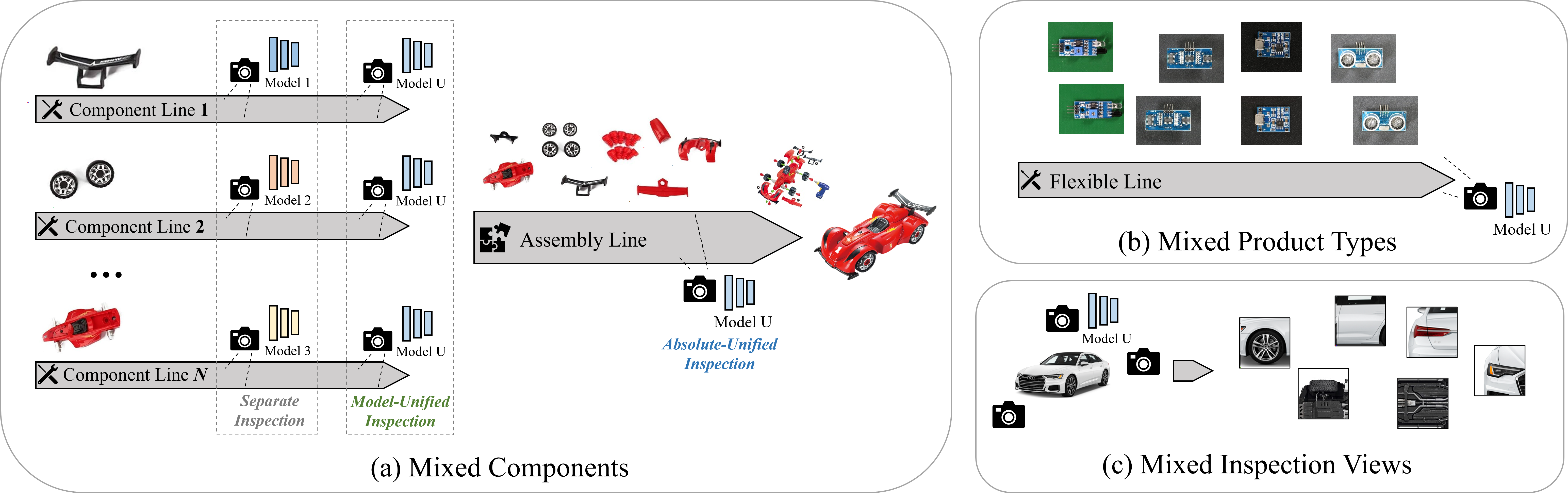}}
\caption{Application scenarios of \textit{absolute-unified} multi-class UAD.  (a) Components of a product are mixed on the assembly line. (b) A product has various types on a flexible manufacturing line. (c) Mixed inspection views for different parts of a large product.}
\label{fig:2}
\end{figure*}

Nevertheless, though the unified model is optimized by a mixture of all classes regardless of object type, it is evaluated on each class respectively during the test phase with different anomaly decision thresholds (including sliding thresholds in area-under-curve metrics such as AUROC). Notably, this evaluation configuration is equivalent to requiring the object class of each image during inference. Prior methods, including those designed for model-unified setting \cite{you2022unified,guo2023recontrast}, demonstrate unsatisfactory results when evaluated with all classes joined together. Moreover, this separate-test paradigm is actually still uncongenial to the scenarios where the normal samples manifest themselves in a large intra-class diversity (i.e., one object consists of various types), which partially contradicts the initial motivation \cite{you2022unified}. Therefore, such methods do not apply to various application scenarios presented in \cref{fig:2} in which image classes are not directly accessible.

To this end, this paper aims to present an approach to address the aforementioned challenging task from the perspective of intra-normal variance. First, we introduce a new and practical anomaly detection setting, i.e., \textit{absolute-unified} UAD, as shown in \cref{fig:1}c.
This task setting requires learning a unified model that is trained and evaluated on a mixture of multiple categories without class labels (unified-train-unified-test). Second, we target the deficiency of prior methods that produce unfavorable performance on \textit{absolute-unified} setting. The anomaly score distributions differ between object classes due to inconsistent intrinsic reconstruction error, resulting in difficulties in obtaining a unified decision threshold, as illustrated in \cref{fig:2}. Accordingly, we propose a simple yet effective method, \textbf{C}lass-\textbf{A}gnostic \textbf{D}istribution \textbf{A}lignment (CADA), to align the anomaly score distributions of different classes to a unification without accessing class labels in both training and inference. The main idea of CADA is to model the anomaly score distribution of normal samples for each implicit class and predict the distribution statistics given any input image of this class. Then, the anomaly map of a test image is normalized by the distribution statistics of its corresponding class so that the normal regions of each class conform to the same distribution (\cref{fig:2}). Furthermore, CADA is a general pipeline that can be integrated into most base UAD methods with little computation overhead.

To validate our effectiveness under \textit{absolute-unified} multi-class UAD setting, we evaluate CADA with extensive experiments on two widely used industrial defect detection benchmarks, i.e., MVTec AD \cite{bergmann2019mvtec} and VisA \cite{zou2022spot}, where our proposed method significantly boosts the performances of conventional approaches. Notably, we present an excellent image-level AUROC of 98.6\% on MVTec AD, which exceeds the previous state-of-the-art (SOTA) of 91.4\% by a large margin.

\section{Related Work}
\label{sec:related}

\subsection{Unsupervised Anomaly Detection}

\textbf{Epistemic methods} are based on a reasonable assumption that the networks respond differently during inference between seen input and unseen input. Specifically, during the inference process, the network cannot correctly reconstruct the anomaly region as it has never been seen during the training since anomaly data are beyond the cognitive space formed by the anomaly-free training data. Within this paradigm, \textbf{pixel reconstruction} methods adopt self-reconstruction models such as autoencoder (AE) \cite{collin2021improved,perera2019learning}, variational autoencoders (VAE) \cite{kingma2013auto,milkovic2021ultrasound}, and generative adversarial network (GAN) \cite{schlegl2019f,zhao2021anomaly,akcay2019ganomaly}. Recently, diffusion models \cite{zhang2023diffusionad,wyatt2022anoddpm} have been used in anomaly detection tasks of industrial and medical. However, the assumption mentioned above is not entirely valid. When anomaly and anomaly-free data for training share common pixel composition patterns (e.g., local edges), pixel-level reconstruction models can successfully recover unseen anomaly regions \cite{you2022unified}. Therefore, the \textbf{feature reconstruction} \cite{you2022unified,you2022adtr,shi2021unsupervised} and \textbf{knowledge distillation} methods \cite{Deng2022,salehi2021multiresolution,wang2021student} is proposed to reconstruct the features of a pre-trained encoder rather than raw pixels. Following the assumption mentioned above, the goal is to mimic the features generated by the teacher network when given input images without anomalies.\\
\textbf{Pseudo-anomaly} methods typically involve generating anomalies on anomaly-free images artificially, transforming unsupervised AD tasks into supervised classification \cite{li2021cutpaste} or segmentation \cite{zavrtanik2021draem} tasks. CutPaste \cite{li2021cutpaste} cuts sections from anomaly-free images and pasts them at arbitrary locations within these images to simulate anomaly generation. DRAEM \cite{zavrtanik2021draem} uses Perlin noise as the mask and adds another image source to construct anomaly regions. SimpleNet \cite{liu2023simplenet} introduces Gaussian noise to generate anomaly features in pretrain feature space. These methods rely heavily on the alignment between the generated pseudo-anomalies and real anomalies, so they may struggle to perform well in complex and diverse anomaly datasets.\\
\textbf{Feature-statistics} methods \cite{defard2021padim,roth2022towards,rippel2021modeling,lee2022cfa} follow a simple assumption that the features of anomaly-free images are distinct from those of anomalous images. This paradigm usually utilizes pre-trained networks on large-scale datasets to extract features from all the normal training samples. During inference, feature memory is used to compare with the test samples in the feature space to detect anomaly targets. PaDiM \cite{defard2021padim} uses a network pre-trained on ImageNet to extract features of anomalous patches through a multivariate Gaussian distribution embedding. PatchCore \cite{roth2022towards} also builds on the same pre-trained network and uses the most representative nominal patch features in a memory bank. During testing, the input features are scored using the maximum feature distance. Due to the need to remember, process, and match key features from the training samples, these methods are computationally intensive during both training and inference, especially when the training dataset exhibits high diversity.

\subsection{Multi-Class Anomaly Detection}

UniAD \cite{you2022unified} first introduced multi-class anomaly detection, aiming to detect anomalies for different classes using a unified model. In this setting, conventional UAD methods often face the challenge of "identical shortcuts", where both anomaly-free and anomaly samples can be effectively recovered during inference \cite{you2022unified}. It is caused by the diversity of multi-class normal patterns that drive the network to generalize on unseen patterns. This contradicts the fundamental assumption of epistemic methods. Most current researches focus on addressing this challenge \cite{you2022unified,lu2023hierarchical,guo2023recontrast,liu2023mixed,yin2023lafite}. For example, UniAD \cite{you2022unified} employs a neighbor-masked attention module and a feature jittering strategy to mitigate these shortcuts. HVQ-Trans \cite{lu2023hierarchical} proposes a VQ-based Transformer model that induces large feature discrepancies for anomalies to solve it. LafitE \cite{yin2023lafite} utilizes a latent diffusion model and introduces a feature editing strategy to alleviate this issue. OmniAL \cite{zhao2023omnial} focuses on anomaly localization in the unified setting, preventing identical reconstruction by synthesizing anomaly data rather than directly using normal data.  It's important to note that all previous arts on multi-class UAD adopt class-separated testing (\cref{fig:1}b). Though there is no explicit statement prohibiting them from unified testing on a mixture of all classes, in our subsequent experiments, we observed a noticeable drop in performances when evaluating these methods under \textit{absolute-unified} setting.

In this work, we focus on \textbf{sensory AD} that detects regional defects (common in practical applications such as industrial inspection and medical disease screening), which is distinguished from \textbf{semantic AD}. In multi-class sensory AD, normal and anomalous samples are the same objects except for local anomaly, e.g. good cable+bottle vs. spoiled cable+bottle. In multi-class semantic AD, the class of normal samples and anomalous samples are semantically different, e.g. cats+dogs vs. cars+trucks. Therefore, Semantic AD is already 'absolute-unified', inferencing with mixed classes (e.g. 5 vs. 5 class setting of CIFAR-10). It does not suffer from absolute-unified setting because anomalous samples do not share the same classes with normal samples.

\section{Method}

The contents of this section are organized as follows. We first discuss why conventional UAD methods designed for \textit{separate} setting and \textit{model-unified} setting suffer from severe degradation when evaluated with all classes mixed together. Second, we present two intuitive solutions to the problem, relaxing the constraint of class label availability, i.e., providing class labels in both training and test, or just in training. Lastly, we introduce CADA as a natural next-step solution when class labels are absolutely unavailable. 

\begin{figure*}[!h]
\centering
\centerline{\includegraphics[width=0.6\textwidth]{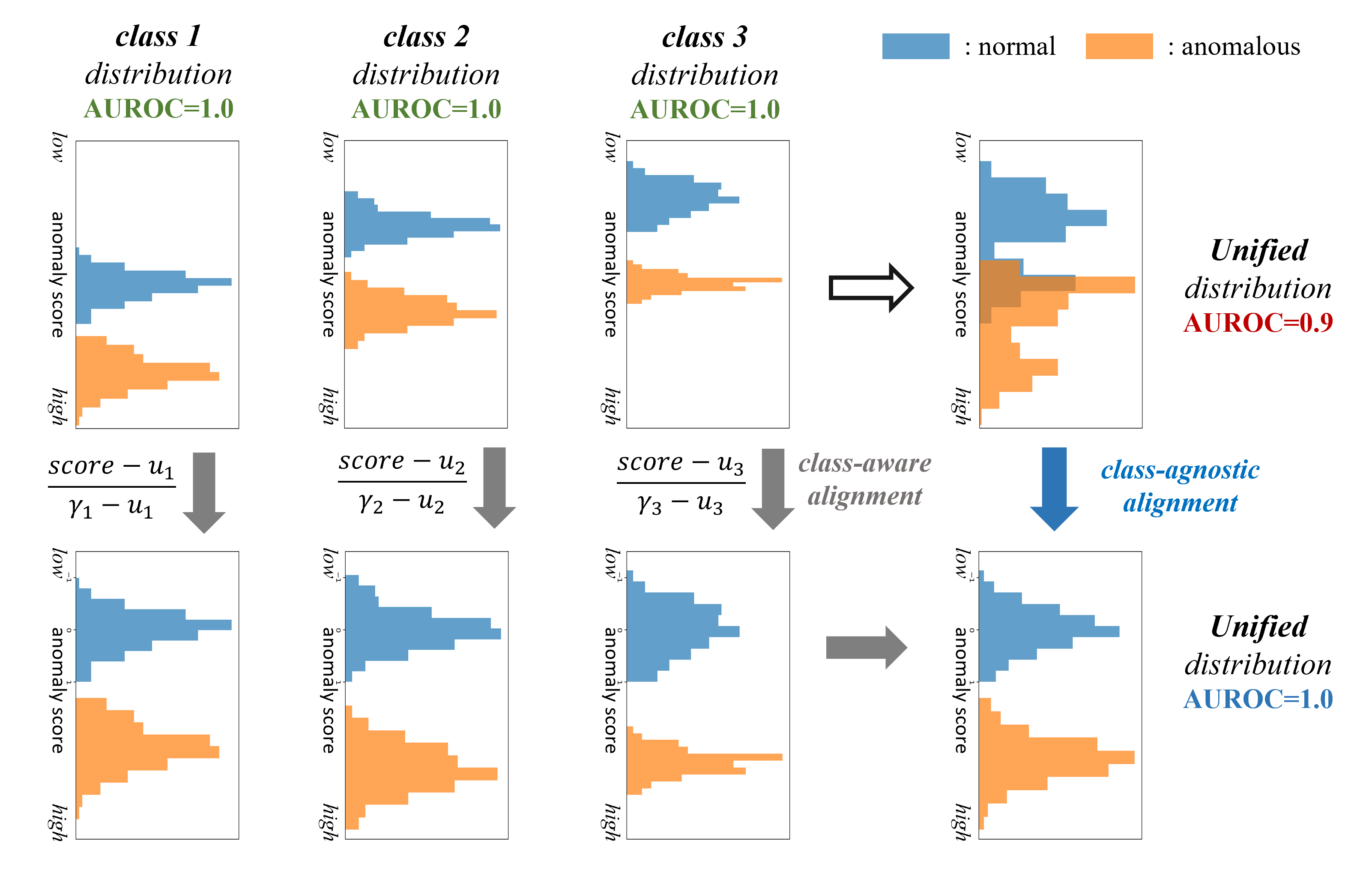}}
\caption{A toy example of why methods fail in \textit{absolute-unified} UAD. First row: Normal and anomalous samples cannot be separated when three classes are mixed. Second row: For each class, the anomaly scores are normalized by the mean (${u}_{c}$) and maximum (${\gamma}_{c}$) of the normal samples; hence, the unified distribution is separable. Under \textit{absolute-unified} setting, ${u}_{c}$ and ${\gamma}_{c}$ cannot be directed computed as images are "anonymous". CADA can estimate and align the distribution without knowing image classes.}
\label{fig:3}
\end{figure*}

\subsection{Preliminary: Inter-Class Distribution Mismatch}
Feature-reconstruction methods \cite{you2022unified,Deng2022,shi2021unsupervised} hold a prominent place in unsupervised anomaly detection. Without loss of generality, we revisit a powerful and widely-used reconstruction-based method, Reverse Distillation (RevDist) \cite{Deng2022}, probing the potential of such methods under \textit{absolute-unifed} setting.  RevDist consists of a frozen pre-trained encoder, a bottleneck, and a decoder. The ImageNet \cite{russakovsky2015imagenet} pre-trained encoder (e.g., ResNet50 \cite{he2016resnet}) extracts informative feature maps with three spatial sizes. The bottleneck is similar to the last layer of the encoder but takes the multi-scale feature maps of the encoder as input. The decoder is the reverse of the encoder, i.e., the down-sampling operation at the beginning of each layer is replaced by up-sampling. During training, the decoder learns to reconstruct the output of the encoder layers by minimizing the cosine distance between feature maps. During inference, the decoder is expected to reconstruct normal regions of feature maps but fail for anomalous regions as it has never seen such samples. The output anomaly map $\mathcal{M}^{map}=\{ s^{pix}_{i,j} \}_{i,j=1}^{H,W} \in \mathbb{R}^{H \times W}$ is given by the per-region feature distances up-sampled to the image size ($H \times W$), where $s^{pix}$ represents pixel-level anomaly score. Commonly, the image-level anomaly score $s^{img}$ of an image is its max $s^{pix}$ \cite{Deng2022, roth2022towards}, or the mean of its $n\%$ highest $s^{pix}$ (e.g., 1\% in our study), representing the most anomalous region. Most feature-reconstruction-based methods adopt very similar paradigms with moderate differences, such as decoder network structure (transformer \cite{you2022adtr,you2022unified}, diffusion models \cite{yin2023lafite}), feature noise \cite{you2022unified}, and trainable encoder \cite{guo2023recontrast}.

In \cref{fig:5}, we visualize image-level and pixel-level anomaly scores $s^{img}$ of normal images \textit{vs.} anomalous images, using RevDist trained on all images of MVTec~AD \cite{bergmann2019mvtec}. It is noticed that the distribution of anomaly scores varies across different categories of images. Due to network information capacity, intrinsic reconstruction errors exist not only in unseen anomalous regions but also in normal regions. The errors of details and edges are generally higher than plain regions. Since different categories of images have different textures, features, and fine-grainedness, their reconstruction difficulties might vary to a large extent, resulting in the variance of inter-class distribution. This distribution mismatch results in greater overlaps between normal and anomaly when jointing all classes to unification, in turn producing unfavorable detection performances, as shown in \cref{fig:3}.

\subsection{Class-Aware Distribution Alignment} \label{sec:3.2}

\begin{figure*}[!t]
\centering
\centerline{\includegraphics[width=\textwidth]{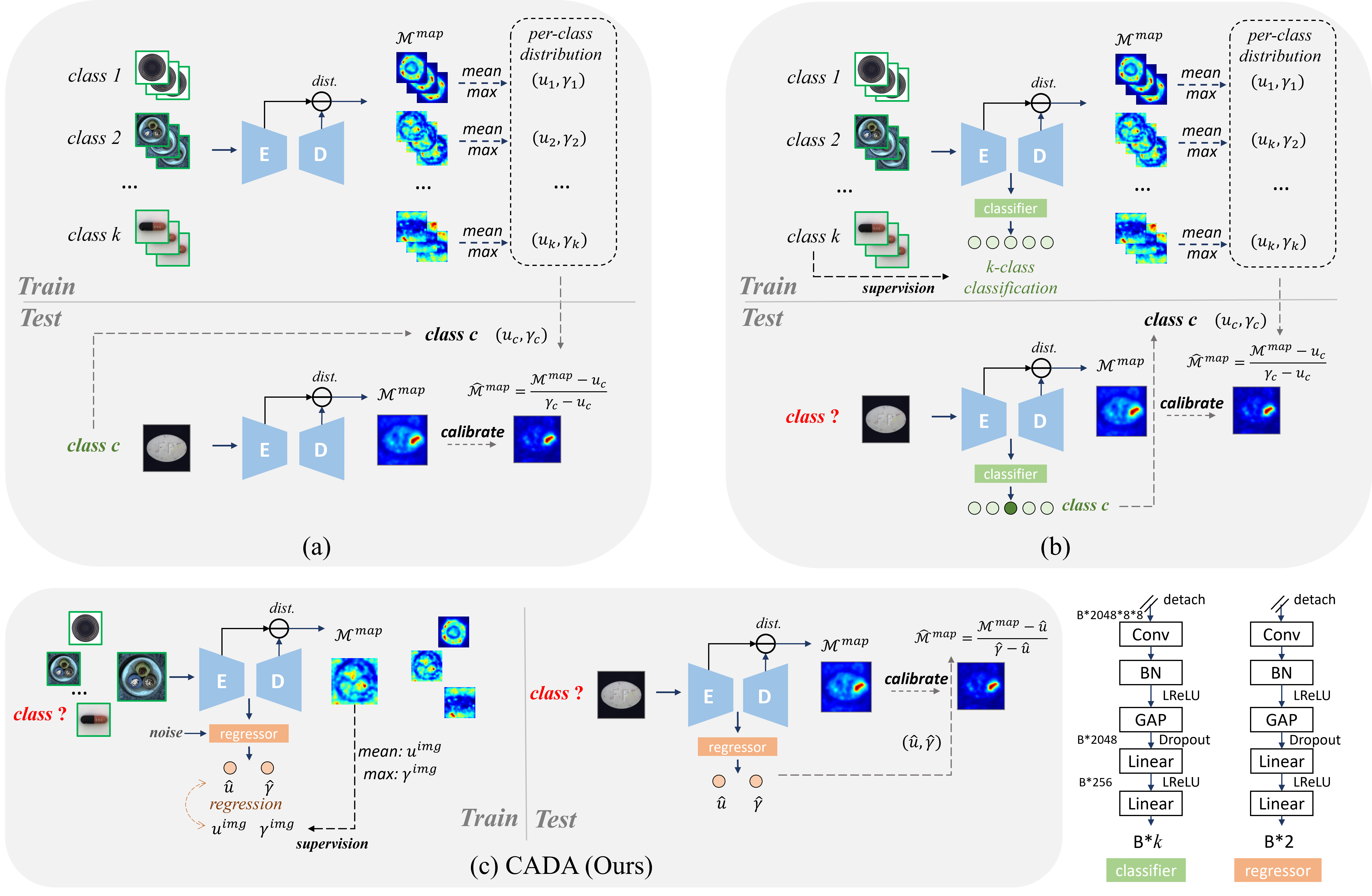}}
\caption{Anomaly score alignment for \textit{absolute-unified} UAD. (a) Class-aware. Class labels are provided in training and test phase. (b) Class-aware. Class labels are provided only in training phase. (c) Class-agnostic. Class label not provided (CADA, Ours). }
\label{fig:4}
\end{figure*}

The aforementioned inter-class distribution mismatch can be easily addressed when class labels are available.

First, we relax the accessibility of class labels in both training and testing, which makes the setting similar to \textit{model-unified} but requires a universal decision threshold for all classes. Apparently, we can derive the per-class anomaly score distributions from the training set, which are used to calibrate the anomaly score of a test image given the class it belongs to, as illustrated in \cref{fig:4}a. After training of the base UAD model, the distributions of normal pixel anomaly score $s^{pix}$ in each class can be measured and described by their expectation (mean) $u_{c}=mean(\{ s^{pix} | x \in \mathcal{X}_c \})$ and maximum $\gamma_{c}=max(\{ s^{pix} | x \in \mathcal{X}_c \})$, where $c \in \{1,...,k\}$ denotes the class number, $x$ denotes an image sample, and $\mathcal{X}_c$ denotes all images of class $c$. In practice, we adopt $\gamma_{c}$ to be the mean of maximum $s^{pix}$ of each image to avoid outliers. Then in test, the output anomaly map $\mathcal{M}^{map}$ of class $c$ is calibrated by mean-max normalization, as:

\begin{equation}
\hat{\mathcal{M}}^{map}=\frac{\mathcal{M}^{map}-u_{c}}{\gamma_{c}-u_{c}}.
\label{eq:1}
\end{equation}
Assuming samples in the training set and normal samples in the test set follow the same distribution, this linear transformation maps normal $s^{pix}$ of each class to the same distribution with zero-mean, one-max, while keeping the correlation between normal and anomaly within a class. As $s^{pix}$ and $s^{img}$ are derived from $\mathcal{M}^{map}$, they go through the same operation. As illustrated in \cref{fig:3}, the joint distributions of multi-classes united are more separable concerning normal and anomaly. Notably, it is not possible to use distributions of the test sets to normalize test anomaly maps, as the output of each test sample is considered to be independent in practice.

Second, we tighten the availability of class labels to the training set only, which is also a worthwhile practical setting. As the class labels of test images are not provided during inference, an intuitive workaround is to build a classification model utilizing the images and labels of the training set to predict class labels for test images. Generally, the classification model can be a standard classification network of any architecture, e.g., ResNet \cite{he2016resnet}, ViT \cite{dosovitskiy2020vit}. Nevertheless, a standalone classification network brings additional computational overhead, which can be redundant in real-time applications \cite{batzner2023efficientad,liu2023simplenet} like industrial inspection. Therefore, we employ a lightweight classification head (classifier) on top of the base UAD network, reusing the middle features, i.e., features from the bottleneck, as shown in \cref{fig:4}b. The structure of the classifier can be flexible, from a single linear layer to multi-layer perception (MLP). Without loss of generality, we adopt one convolution and two linear layers as the classifier by default. The classifier can be optimized by cross-entropy loss, as:
\begin{equation}
L_{CE} = -\sum_{c=1}^ky_{c}\log(p_{c})
\label{eq:2}
\end{equation}
where $p_c$ is the output probability of class $c$, and $y_c=1$ if $x \in \mathcal{X}_c$, otherwise $y_c=0$. During test, the class with the largest $p_c$ draws the corresponding distribution $(u_{c}, \gamma_{c})$, which is used to calibrate $\mathcal{M}^{map}$ as in \cref{eq:1}.

\subsection{Class-Agnostic Distribution Alignment} 
Naturally, we restrict the accessibility of class labels throughout all phases, which follows the \textit{absolute-unified} multi-class UAD setting. In this setting, we propose a simple yet effective method, \textbf{C}lass-\textbf{A}gnostic \textbf{D}istribution \textbf{A}lignment (CADA), to estimate anomaly score distribution statistics, i.e., $(u_{c},\gamma_{c})$, of normal samples of the class that an input test image belongs to. The core of CADA is to estimate the distribution directly from an image, simplifying the distribution estimation paradigm of \cref{fig:4}b that undergoes an inference process of image-to-class-to-distribution which requires class labels. To be specific, a simple regressor, the same structure as the classifier, is trained to regress $(u_{c},\gamma_{c})$ taking features from the pre-trained encoder, as shown in \cref{fig:4}c. Because $u_{c}$ and $\gamma_{c}$ are not available, the mean and maximum $s^{pix}$ of each training image, denoted by $u^{img}$ and $\gamma^{img}$, can serve as a decent approximation, as the normal images in one class follow the same distribution and have similar appearances. During training, $u^{img}$ and $\gamma^{img}$ are used as the ground-truth in regression loss, as:

\begin{equation}
L_{reg}={L1}_{smooth}(\hat{u},u^{img})+{L1}_{smooth}(\hat{\gamma},\gamma^{img}) ,
\label{eq:3}
\end{equation}

\begin{equation}
{L1}_{smooth}=
    \left\{\begin{matrix}
        \frac{1}{2\alpha}(y - \hat{y})^{2} & if \left | y - \hat{y}  \right | < \alpha\\
        \left | y - \hat{y}  \right | - \frac1 2 \alpha & otherwise
    \end{matrix}\right. ,
\label{eq:4}
\end{equation}
where $\hat{u}$ and $\hat{\gamma}$ are outputs of the regressor, $L1_{smooth}$ is smoothed $L_1$ loss widely used for regression tasks.

It is worth noting that CADA is expected to predict the distribution of normal samples no matter the input image is normal or anomalous. Same as previous, $\mathcal{M}^{map}$ is calibrated by mean-max normalization, as:
\begin{equation}
\hat{\mathcal{M}}^{map}=\frac{\mathcal{M}^{map}-\hat{u}}{\hat{\gamma}-\hat{u}}.
\label{eq:5}
\end{equation}
We expect that the regressor can perform the inference process of image-to-class-to-distribution like in \cref{fig:4}b, instead of simply giving the $u^{img}$ and $\gamma^{img}$ of each test image. However, the regressor could have generalized too well on anomalous samples, predicting exactly the mean and maximum $s^{pix}$ of an anomalous anomaly map, which in turn loses the effect of anomaly score calibration.

To tackle the above issues, we add random noise into the regressor during training, without influencing the base UAD network. A simple Dropout \cite{hinton2012dropout} before linear layers can work well enough. Trivial as it is, the injected noise can be explained from three perspectives. On one hand, these noises can be regarded as pseudo anomalies. The regressor is trained to regress the distribution of normal samples, even though its features contain anomalies. On the other hand, it serves as a regularization for the hidden classification task. Because anomalous images are not involved during training, such noises help the network to generalize its classification ability from available normal images to unseen anomalous images. In addition, the random perturbation prevents the network from overfitting to $(u^{img}, \gamma^{img})$ of each image that serves as the noisy approximation of $(u_{c},\gamma_{c})$.

\section{Experiment}

\subsection{Datasets and Metrics}

\textbf{MVTec AD} \cite{bergmann2019mvtec} is an industrial defect detection dataset, containing 15 sub-datasets (5 texture classes and 10 object classes) with a total of 3,629 normal images as the training set and 1,725 images as the test set (467 normal, 1258 anomalous). Pixel-level annotations are available for anomalous images for evaluating anomaly segmentation. All images are resized to 256$\times$256.

\textbf{VisA} \cite{zou2022spot} is an industrial defect detection dataset, containing 12 sub-datasets with a total of 9,621 normal and 1,200 anomalous images. Training and test sets are split following the official setting \cite{zou2022spot}, resulting in 8659 images in the training set and 2162 images in the test set (962 normal, 1,200 anomalous).

\textbf{Metrics}.
Image-level anomaly detection performance is measured by Area Under the Receiver Operator Curve (I-AUROC) and Area Under the Precision Recall Curve (I-AP).  Pixel-level anomaly segmentation (localization) is measured by pixel-level AUROC (P-AUROC) and pixel-level AP (P-AP). AP is more meaningful than AUROC for evaluating anomaly segmentation due to the unbalanced amount of normal and anomalous pixels \cite{zavrtanik2021draem,zou2022spot}.

\subsection{Implementation Details}

We integrate CADA on three base UAD methods: UniAD \cite{you2022unified}, RevDist \cite{Deng2022}, and ReContrast \cite{guo2023recontrast}. Other reproduced methods \cite{zavrtanik2021draem,roth2022towards,lee2022cfa} yield unconvincing results even under \textit{model-unified} setting, so we leave them unconsidered. For RevDist and ReContrast, WideResNet50 \cite{zagoruyko2016wide} is used as the encoder following the original implementation. GELU substitutes the ReLU activation in the decoder for better training stability. The input image size is 256$\times$256. AdamW optimizer \cite{loshchilov2018adamw} is utilized with learning rate(lr)=2e-3, $\beta$=(0.9,0.999) and weight decay(wd)=1e-5. The network is trained for 5,000 iterations with batch size of 16 on each dataset. For UniAD, EfficientNet-B4 is used following their implementation. The image size is 224$\times$224.  AdamW optimizer \cite{loshchilov2018adamw} is utilized with lr=3e-4, $\beta$=(0.9,0.999) and wd=1e-4. The network is trained for 50,000 iterations with batch size of 64 on each dataset.

After the training of base UAD methods, the CADA regressors of RevDist, ReContrast, and UniAD are optimized with the batch size of 16 for 5,000, 5,000, and 20,000 iterations, respectively. SGD optimizer is adopted with lr=5e-2, momentum=0.9, and wd=1e-4. Smoothing thresh $\alpha$ in ${L1}_{smooth}$ is set to 0.1, 0.1, and 10 for RevDist, ReContrast, and UniAD respectively according to the scale of their anomaly scores. The regressor of CADA employs the convolution layer with the same output dimension as the input (bottleneck feature). The hidden dimension of the linear layers is 256.  The dropout rate is 0.25 by default. The image-level anomaly score $s^{img}$ is the mean of its $1\%$ highest $s^{pix}$. With or without CADA, these base methods follow the same hyperparameters for fair comparison. Codes are implemented with Python 3.8 and PyTorch 1.12.0, and will be available upon acceptance. Experiments are run on NVIDIA GeForce RTX3090 GPUs (24GB).

\begin{table*}[!t]
\tiny
\caption{Anomaly detection and localization performance (\%) on MVTec AD, measured in I-AUROC / I-AP and P-AUROC / P-AP, respectively.}
\label{tab:1}
\centering
\begin{tabular}{@{}lll|ccccc|ccc@{}}
\toprule
 & setting & classes      & DRAEM & PatchCore & UniAD & RevDist & ReContrast & \makecell[c]{UniAD\\+CADA} & \makecell[c]{RevDist\\+CADA} & \makecell[c]{ReContrast\\+CADA} \\ \midrule
 
\multirow{2}{*}{\rotatebox{90}{\textbf{Detect.}}} & \makecell[c]{\textit{Model} \\ \textit{Unified}} & 15 avg.    &       89.1/94.7&           96.3/99.0&       97.1/99.0&         97.8/99.1& 98.5/99.4           &            97.2/99.1 &              98.2/99.2&   98.7/99.5    \\ \cmidrule(l){2-11}

&  \makecell[c]{\textit{\textbf{Absolute}} \\ \textit{\textbf{Unified}}}  & \textbf{mixed}           &       82.3/93.7&           89.0/96.2&       91.4/97.0&         85.0/94.4 &   90.7/96.8         &            95.7/98.4 &              98.0/99.3&      \textbf{98.6}/\textbf{99.5}  \\ 

& &           &       &      &    &  \multicolumn{2}{c}{$\Delta$ w.r.t. the base:}   &  \textcolor{Green}{\textbf{+4.3}}/\textcolor{Green}{\textbf{+1.4}}    & \textcolor{Green}{\textbf{+13.0}}/\textcolor{Green}{\textbf{+4.9}} &     \textcolor{Green}{\textbf{+7.9}}/\textcolor{Green}{\textbf{+2.7}}  \\ \midrule \midrule

\multirow{2}{*}{\rotatebox{90}{\textbf{Local.}}} & \makecell[c]{\textit{Model} \\ \textit{Unified}} & 15 avg.    &       88.9/52.2&           75.4/20.3&       97.1/53.1&         96.4/58.1& 97.6/60.8&            97.0/52.7 &              96.1/57.0&   97.8/61.3\\ \cmidrule(l){2-11}

&  \makecell[c]{\textit{\textbf{Absolute}} \\ \textit{\textbf{Unified}}}  & \textbf{mixed}           &       81.8/41.1&           84.4/25.6&       96.4/51.1&         94.5/49.0 &   96.8/56.3 &            96.1/50.9&              95.6/53.1&      \textbf{97.2}/\textbf{57.2}\\ 

& &           &       &      &    &  \multicolumn{2}{c}{$\Delta$ w.r.t. the base:}   &  -0.3/-0.2 & \textcolor{Green}{\textbf{+1.1}}/\textcolor{Green}{\textbf{+4.1}}&     \textcolor{Green}{\textbf{+0.4}}/\textcolor{Green}{\textbf{+0.9}}  \\ \bottomrule

\end{tabular}
\end{table*}

\begin{table*}[!t]
\tiny
\caption{Anomaly detection and localization performance (\%) on VisA, measured in I-AUROC / I-AP and P-AUROC / P-AP, respectively.}
\label{tab:2}
\centering
\begin{tabular}{@{}lll|ccccc|ccc@{}}
\toprule
& setting & classes      & DRAEM & CFA& UniAD & RevDist & ReContrast & \makecell[c]{UniAD\\+CADA} & \makecell[c]{RevDist\\+CADA} & \makecell[c]{ReContrast\\+CADA} \\ \midrule

\multirow{2}{*}{\rotatebox{90}{\textbf{Detect.}}} & \makecell[c]{\textit{Model} \\ \textit{unified}} & 12 avg.    &       75.1/78.9&           74.1/78.0&       90.9/93.4&         93.8/94.8 & 95.8/96.7&            90.5/92.7   &              93.4/94.8&   95.8/96.7\\ \cmidrule(l){2-11} 
& \makecell[c]{\textit{\textbf{Absolute}} \\ \textit{\textbf{Unified}}}  &\textbf{mixed}     &       70.0/76.0&           42.9/50.1&       89.0/91.8&         83.5/88.2 &   93.2/95.0&            90.7/93.3 &              92.8/94.4&      \textbf{95.8}/\textbf{96.7}\\ 

& &           &       &      &    &  \multicolumn{2}{c}{$\Delta$ w.r.t. the base:}   &  \textcolor{Green}{\textbf{+1.7}}/\textcolor{Green}{\textbf{+1.5}}& \textcolor{Green}{\textbf{+9.3}}/\textcolor{Green}{\textbf{+6.2}}&     \textcolor{Green}{\textbf{+2.6}}/\textcolor{Green}{\textbf{+1.7}}\\ \midrule \midrule

\multirow{2}{*}{\rotatebox{90}{\textbf{Local.}}} &\makecell[c]{\textit{Model} \\ \textit{unified}} & 12 avg.    &       87.1/10.3&           94.0/33.1&       98.3/40.4&         96.3/44.2& 98.5/48.4&            98.2/39.8 &              96.2/44.5&   98.3/48.1\\ \cmidrule(l){2-11} 
& \makecell[c]{\textit{\textbf{Absolute}} \\ \textit{\textbf{Unified}}}  & \textbf{mixed}           &       82.7/7.0&           72.9/1.6&       97.9/34.7&         94.0/32.9&   97.1/42.1&            \textbf{98.3}/35.4&              95.4/41.7&      97.8/\textbf{45.0}\\ 

& &           &       &      &    &  \multicolumn{2}{c}{$\Delta$ w.r.t. the base:}   &  \textcolor{Green}{\textbf{+0.4}}/\textcolor{Green}{\textbf{+0.7}} & \textcolor{Green}{\textbf{+1.4}}/\textcolor{Green}{\textbf{+8.8}}&     \textcolor{Green}{\textbf{+0.7}}/\textcolor{Green}{\textbf{+2.9}}\\ \bottomrule
 
\end{tabular}
\end{table*}

\subsection{Main Results}

\begin{figure*}[!t]
\centering
\centerline{\includegraphics[width=\textwidth]{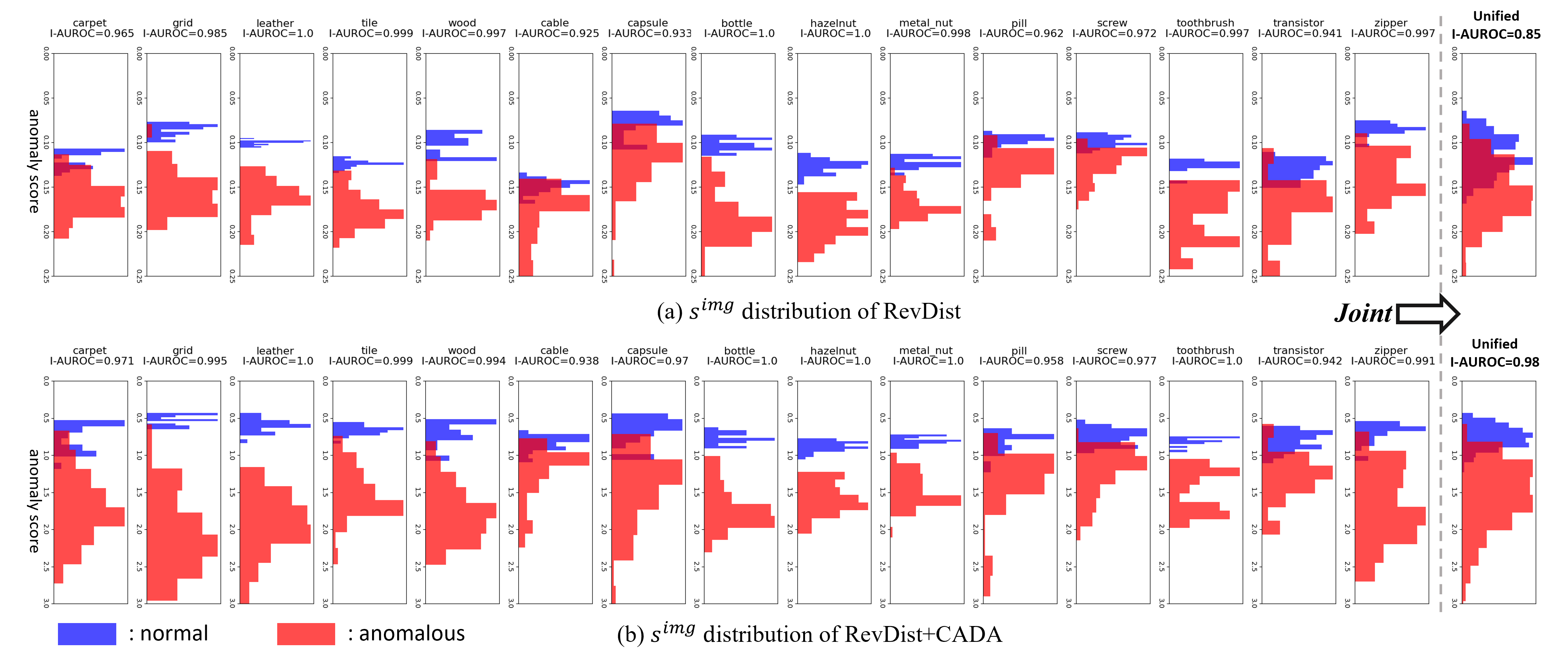}}
\caption{Image-level anomaly score distribution on MVTec AD. (a) RevDist. (b) RevDist+CADA. The distributions of normal samples are more consistent across different classes with the proposed CADA.}
\label{fig:5}
\end{figure*}

\begin{figure}[!t]
\centering
\centerline{\includegraphics[width=0.6\textwidth]{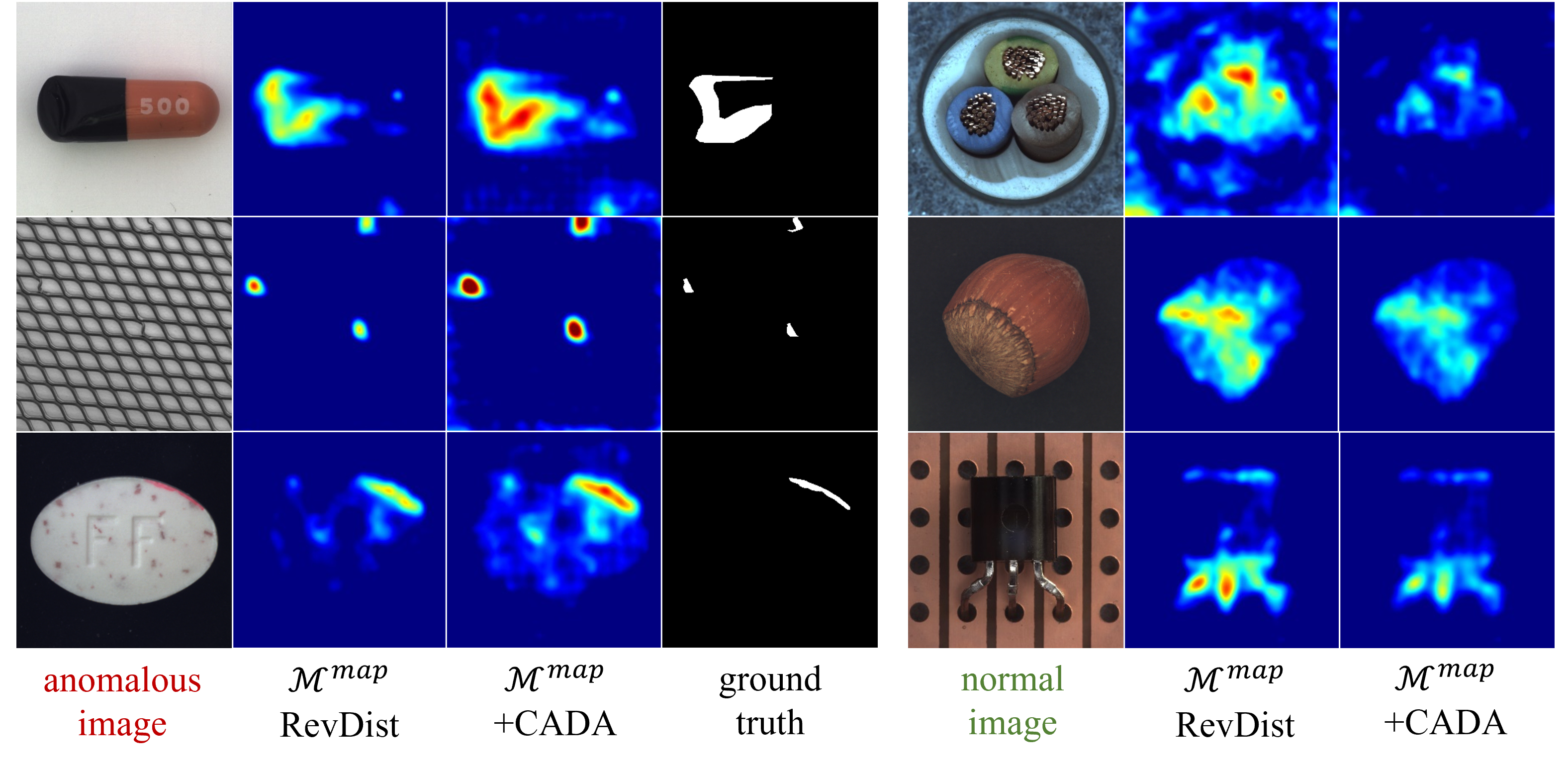}}
\caption{Qualitative results of RevDist and RevDist+CADA. $\mathcal{M}^{map}$ is visualized by mapping dataset-average image mean and image maximum $s^{pix}$ to 0 and 1. }
\label{fig:6}
\end{figure}

\textbf{MVTec AD} \cite{bergmann2019mvtec}.  We reproduce various base UAD methods under \textit{absolute-unified} multi-class setting on MVTec~AD \cite{bergmann2019mvtec}, including DRAEM \cite{zavrtanik2021draem}, PatchCore \cite{roth2022towards}, UniAD \cite{you2022unified}, RevDist \cite{Deng2022}, and ReContrast \cite{guo2023recontrast}. We also report their performances under \textit{model-unified} multi-class setting as comparisons. Image and pixel-level performances are reported in \cref{tab:1}. Compared with UniAD, the previous SOTA, our method (CADA+ReContrast) significantly improves I-AUROC from 91.4\% to 98.6\%, by a large margin of \textbf{7.2\%}. Compared with base UAD methods, integrating CADA boosts these methods to the next level, increasing the I-AUROC of UniAD, RevDist, and ReContrast by 7.9\%, 13.0\%, and 4.3\%, respectively. For anomaly segmentation, we also achieve SOTA P-AUROC of 97.2\%  and P-AP of 57.2 \%. Our method slightly harms the segmentation performance of UniAD, which may be caused by the noise introduced by feature jitter during training. The per-class $s^{img}$ distribution is presented in \cref{fig:5}, demonstrating that our method can align the mismatched score distribution of RevDist. The distribution of $s^{pix}$ is presented in Appendix. Qualitative results of anomaly localization are presented in \cref{fig:6}, where CADA produces better localization on anomalous images and less activation on normal images.

\textbf{VisA} \cite{zou2022spot}. The experimental settings are the same as MVTec~AD, except that PatchCore \cite{roth2022towards} cannot be reproduced on the much larger training set due to its high CPU memory consumption. To compensate for this lack, we reproduce a similar method, CFA \cite{lee2022cfa}. Image and pixel-level performances are reported in \cref{tab:2}. The performance gap between \textit{model-unified} and \textit{absolute-unified} setting is smaller than that on MVTec~AD, as VisA consists of similar categories, e.g. four categories are PCB. Among the base methods above, ReContrast achieves the best performance. Our approach (ReContrast+CADA) further improves anomaly detection performance, increasing I-AUROC from 93.2\% to 95.8\%. Our method also shows powerful ability for other base methods. Compared with RevDist with unsatisfactory performance, CADA significantly improves I-AUROC from 83.9\% to 95.4\%, by a large \textbf{11.5\%}. For anomaly segmentation, our method also shows improvement over previous methods that have worked well, achieving SOTA P-AUROC of 97.8\% and P-AP of 45.0\%. The improvement of integrating CADA on UniAD is not obvious, as UniAD demonstrates limited performance on VisA under \textit{model-unified} setting.

\begin{table*}[!t]
\tiny
\caption{Ablations of regressor structure, dropout rate, and $s^{img}$, conducted on MVTec~AD (\%).}
\label{tab:3}
\centering
\begin{tabular}{@{}l|c|c|c|cccc@{}}
\toprule
          & \textit{regressor structure} & \textit{dropout rate} & \textit{$n\%$ highest  $s^{pix}$ as $s^{img}$} & I-AUROC & I-AP & P-AUROC & P-AP \\ \midrule\midrule
RevDist   & -                   & -        & 1\%    & 85.0    & 94.4 & 94.5    & 49.0 \\ \midrule
~+CADA     & 1 linear            & 0.25    & 1\%      & 96.0    & 98.6 & 95.6    & 53.8 \\
          & 2 linear            & 0.25     & 1\%     & 96.9    & 98.9 & 95.6    & 52.6 \\
          & 3 linear            & 0.25     & 1\%     & 97.1    & 99.0 & 95.3    & 52.4 \\
~~(default) & 1 conv, 2 linear     & 0.25   & 1\%       & \textbf{98.0}& \textbf{99.3} & 95.6 & 53.1\\
          & 2 conv, 2 linear     & 0.25   & 1\%       & 97.3    & 99.1 & \textbf{96.0}    & \textbf{54.8} \\ \midrule
          & 1 conv, 2 linear     & 0      & 1\%       & 97.5    & 99.1 & 95.4    & 52.0 \\
~~(default) & 1 conv, 2 linear     & 0.25  & 1\%        & \textbf{98.0}& \textbf{99.3} & 95.6 & 53.1\\
          & 1 conv, 2 linear     & 0.50   & 1\%        & 97.7    & 99.2 & 95.6    & \textbf{53.5} \\
       & 1 conv, 2 linear     & 0.75     & 1\%      & 97.6    & 99.2 & \textbf{95.8}    & 53.3 \\ \midrule
 & 1 conv, 2 linear     & 0.25  & max $s^{pix}$   & 97.3& 99.0 & 95.6 & 53.1\\
  & 1 conv, 2 linear     & 0.25  & 0.1\%   & 97.8 & 99.2 & 95.6 & 53.1\\
~~(default) & 1 conv, 2 linear     & 0.25  & 1\%   & \textbf{98.0}& \textbf{99.3} & 95.6 & 53.1\\
  & 1 conv, 2 linear     & 0.25  & 2\%   & 97.4 & 99.1 & 95.6 & 53.1\\ \bottomrule

\end{tabular}
\end{table*}

\subsection{Ablation Study}
\textbf{Micro design.} We conduct experiments to investigate the impact of regressor structures, as shown in \cref{tab:3}. The hidden dimension of linear layers is kept to be 256. The results suggest that our CADA is robust to regressor structure, constantly exceeding the baseline RevDist by a large margin. The one-layer regressor produces sub-optimal performance, indicating that a single linear layer is not capable of executing the whole image(feature)-class-distribution inference process. We investigate the impact of Dropout in regressor. As shown in \cref{tab:3}, the dropout slightly boosts the performance by injecting noises to improve the regressor's robustness to the unseen anomalous images not presented in the training. In addition, CADA is robust to different dropout rates.

\textbf{Max pixel or pixels.} Most previous works adopt the maximum $s^{pix}$ of an image as its anomaly score. Although it is not our focus, we demonstrate that using the average of the highest several pixels yields better performance, as shown in \cref{tab:3}.

\textbf{Class-aware or agnostic.} We conduct experiments to compare CADA with its predecessors discussed in \cref{sec:3.2} in which class labels are available to some extent. For CaDA-A (\cref{fig:4}a), the anomaly map is calibrated by $(u_c,\gamma_c)$, given image class labels in all phases. For CaDA-B (\cref{fig:4}b), a classifier with the same structure as the regressor is trained during training to predict the class label for each test image. The results are presented in \cref{tab:5}, where all aligning variants improve the base RevDist. The results of CaDA-A and CaDA-B are identical as the supervised classification task is simple given class information during training. In addition, CADA exceeds CaDA-A and CaDA-B on MVTec-AD, which can be seen as the upper-bound baseline. This phenomenon can be attributed to that CADA further manifests intra-class variance and mismatch beyond pre-defined class labels.

\begin{table}[!t]
\scriptsize
\caption{Comparison of CADA variants, conducted on MVTec~AD (\%).}
\label{tab:5}
\centering
\begin{tabular}{@{}l|c|ccc|ccc@{}}
\toprule
\multirow{2}{*}{} & \multirow{2}{*}{\textit{class info.}} & \multicolumn{3}{c}{MVTec AD} & \multicolumn{3}{c}{VisA} \\
     &   & I-AUROC & P-AUROC & P-AP & I-AUROC & P-AUROC & P-AP\\ \midrule
RevDist     & \textbf{none}            & 85.0   & 94.5    & 49.0  & 83.5 & 94.0 & 32.9 \\ \midrule
+CaDA-A (\cref{fig:4}a)      & train \& inference & 97.3    & 95.2    & 52.4 & 93.0 & 95.6  & 40.3 \\
+CaDA-B (\cref{fig:4}b)     & train       & 97.3         &   95.2      &   52.4 & \textbf{93.0} & \textbf{95.6}  & 40.3  \\
+CADA        & \textbf{none}            & \textbf{98.0}   &\textbf{96.0}    & \textbf{54.7} & 92.8 & 95.4&  \textbf{41.7} \\ \bottomrule
\end{tabular}
\end{table}

\textbf{Estimated statistics.}
A simple variation to describe the distribution is to measure $u_c$ and the standard deviation $\sigma_c=std(\{ s^{pix} | x \in \mathcal{X}_c \})$, substituting $\gamma^{img}$ by $u^{img}+3\cdot\sigma^{img}$ in \cref{eq:3}, and in turn, $\hat{\gamma}$ by $\hat{u}+3\cdot\hat{\sigma}$ in \cref{eq:5}. This variant can be more robust when the normal training set contains noises (anomalous images) because maximum is more sensitive to outliers than standard deviation. We evaluate this variant on MVTec AD, VisA, and noisy BTAD. BTAD \cite{mishra2021btad} is an industrial defect detection dataset comprising 2,830 images of 3 industrial products. This dataset is considered to be noisy, as there are a few anomalous images in the training set \cite{jiang2022softpatch}. The results are presented in \cref{tab:6}. This variant shows robustness to data noise while producing sub-optimal results on the clean MVTec AD but still exceeding the baseline. Adaptations under noisy data can be further investigated.

\begin{table}[!t]
\scriptsize
\caption{Comparison with CADA variant that estimates standard deviation, conducted on MVTec~AD, VisA, and BTAD (\%).}
\label{tab:6}
\centering
\begin{tabular}{@{}ll|c|cccc@{}}
\toprule
dataset            &         & \textit{estimate} & I-AUROC & I-AP & P-AUROC & P-AP \\  \midrule
\multirow{3}{*}{MVTec AD} & RevDist & -          & 85.0          & 94.4          & 94.5          & 49.0          \\
                          & +CADA   & $(u_c,\gamma_c)$      & \textbf{98.0} & \textbf{99.3} & \textbf{96.0} & \textbf{54.7} \\
                          & +CADA   & $(u_c,\sigma_c)$      & 96.3          & 98.7          & 95.3          & 48.2          \\ \midrule
                          
\multirow{3}{*}{VisA} & RevDist & -          & 83.5& 88.2& 94.0& 32.9\\
                          & +CADA   & $(u_c,\gamma_c)$      & \textbf{92.8}& \textbf{94.4}& \textbf{95.4}& \textbf{41.7}\\
                          & +CADA   & $(u_c,\sigma_c)$      & 87.1& 90.9& 95.5& 37.7\\ \midrule
                          
\multirow{3}{*}{\makecell[l]{\textit{noisy}\\BTAD}}     & RevDist & -          & 81.1          & 83.5          & 97.4          & \textbf{60.8 }         \\
                          & +CADA   & $(u_c,\gamma_c)$      &   88.1   & 88.6   &  97.2    &   54.0           \\
                          & +CADA   & $(u_c,\sigma_c)$      & \textbf{98.4}    & \textbf{97.6 }         & \textbf{97.6}        & 51.4          \\ \bottomrule
\end{tabular}
\end{table}

\section{Conclusion}
In this study, we propose a simple yet effective method, CADA, for a novel UAD setting incorporating anonymous multi-class images. It handles the mismatched anomaly score distribution of various classes by aligning them to unification without any class information. The essence of CADA is to predict the distribution statistics of anomaly scores by training a regressor using approximation. Extensive experiments on MVTec AD and VisA demonstrate our superiority, indicating CADA, as a general module, can activate the potential of conventional UAD methods under this challenging setting. We hope the proposed \textit{absolute-unified} multi-class paradigm can further boost the application of anomaly detection methods in a broader range of scenarios.

%
%
\bibliographystyle{splncs04}
\bibliography{egbib}
\end{document}